\documentclass[conference]{IEEEtran}
\IEEEoverridecommandlockouts
\pdfoutput=1

\usepackage{scalerel}
\usepackage{tikz}
\usetikzlibrary{svg.path}

\definecolor{orcidlogocol}{HTML}{A6CE39}
\tikzset{
  orcidlogo/.pic={
    \fill[orcidlogocol] svg{M256,128c0,70.7-57.3,128-128,128C57.3,256,0,198.7,0,128C0,57.3,57.3,0,128,0C198.7,0,256,57.3,256,128z};
    \fill[white] svg{M86.3,186.2H70.9V79.1h15.4v48.4V186.2z}
                 svg{M108.9,79.1h41.6c39.6,0,57,28.3,57,53.6c0,27.5-21.5,53.6-56.8,53.6h-41.8V79.1z M124.3,172.4h24.5c34.9,0,42.9-26.5,42.9-39.7c0-21.5-13.7-39.7-43.7-39.7h-23.7V172.4z}
                 svg{M88.7,56.8c0,5.5-4.5,10.1-10.1,10.1c-5.6,0-10.1-4.6-10.1-10.1c0-5.6,4.5-10.1,10.1-10.1C84.2,46.7,88.7,51.3,88.7,56.8z};
  }
}

\newcommand\orcidicon[1]{\href{https://orcid.org/#1}{\mbox{\scalerel*{
\begin{tikzpicture}[yscale=-1,transform shape]
\pic{orcidlogo};
\end{tikzpicture}
}{|}}}}

\usepackage{tabularx,booktabs}
\newcolumntype{C}{>{\centering\arraybackslash}X} 
\usepackage{mathtools}
\usepackage{amsmath,amssymb,amsfonts}
\usepackage{dsfont}
\usepackage{algorithmic}
\usepackage{graphicx}
\usepackage[backend=bibtex, natbib=true, style=ieee, sorting=none]{biblatex}
\ifCLASSOPTIONcompsoc
    \usepackage[caption=false, font=normalsize, labelfont=sf, textfont=sf]{subfig}
\else
\usepackage[caption=false, font=footnotesize]{subfig}
\fi
\usepackage[hidelinks]{hyperref}
\usepackage{enumerate}
\usepackage{textcomp}
\usepackage{xcolor}
\usepackage{cleveref}
\def\BibTeX{{\rm B\kern-.05em{\sc i\kern-.025em b}\kern-.08em
    T\kern-.1667em\lower.7ex\hbox{E}\kern-.125emX}}
    
\makeatletter
\newcommand{\linebreakand}{%
  \end{@IEEEauthorhalign}
  \hfill\mbox{}\par
  \mbox{}\hfill\begin{@IEEEauthorhalign}
}
\makeatother

\begin{document}

\title{Finetuning BERT on Partially Annotated \\ NER Corpora
\thanks{This work was supported by RFBR Grant 20-07-00561 A.}
}

\author{
    \IEEEauthorblockN{Viktor Scherbakov \IEEEauthorrefmark{1} \IEEEauthorrefmark{2} \orcidicon{0000-0003-1098-9548}}
    \IEEEauthorblockA{viktor.sch@ispras.ru}
    \and 
    \IEEEauthorblockN{Vladimir Mayorov \IEEEauthorrefmark{1} \orcidicon{0000-0003-3882-1114}}
    \IEEEauthorblockA{vmayorov@ispras.ru}
    \linebreakand 
    \IEEEauthorrefmark{1}Ivannikov Institute for System Programming of the Russian Academy of Sciences
    \linebreakand 
    \IEEEauthorrefmark{2}Lomonosov Moscow State University
}

\maketitle

\begin{abstract}
Most Named Entity Recognition (NER) models operate under the assumption that training datasets are fully labelled. While it is valid for established datasets like CoNLL 2003 and OntoNotes, sometimes it is not feasible to obtain the complete dataset annotation. These situations may occur, for instance, after selective annotation of entities for cost reduction. This work presents an approach to finetuning BERT on such partially labelled datasets using self-supervision and label preprocessing. Our approach outperforms the previous LSTM-based label preprocessing baseline, significantly improving the performance on poorly labelled datasets. We demonstrate that following our approach while finetuning RoBERTa on CoNLL 2003 dataset with only 10\% of total entities labelled is enough to reach the performance of the baseline trained on the same dataset with 50\% of the entities labelled.
\end{abstract}


\section{Introduction}

The goal of Named Entity Recognition (NER) is to find mentions of real-world entities from the set of fixed categories in unstructured text. While entity categories could overlap \cite{ultra-fine-ner} and be organized in complex hierarchies \cite{fine-ner}, we will consider the categories to not overlap for the sake of simplicity. Generally, the solutions are based on some ML sequence-to-sequence classification model \cite{hmm-ner, blstm-crf, cnn-lstm-crf, bert}. Each text token (word, punctuation symbol, whitespace, etc., depending on the tokenization used) is assigned a label that carries information about whether this token belongs to an entity of a specific category. By default, every token is labelled as ``Outside of an entity``.

Even though NER is a relatively well-studied field with many high-quality datasets \cite{conll, ontonotes, wnut}, there are still limitations to its application in practice. Since ML models are heavily used, a lot of training data is required, and it is often the case that existing datasets cannot satisfy this requirement due to a mismatch of the category sets. For example, suppose there is an extensive dataset with only a LOCATION entity category, and we need a model that can detect LOCATION and PERSON categories. In that case, we will not be able to use this dataset to train such NER model since there may be a lot of unlabelled PERSON entities.

There are several ways to combat this issue. The first and the most obvious approach is to relabel the whole dataset to introduce a new entity category. Unfortunately, this is a very expensive process due to the significance of the context in which a particular entity was mentioned. For instance, the word ``Washington`` may be somebody's family name or refer to the US capital. The expenses can be cut down with the help of crowdsourcing the annotations \cite{amazon-turk-crowdsourcing, twitter-crowdsource}. However, not all of the entity categories can be crowdsourced. There are some cases that require expert knowledge to obtain the correct annotations. An example of such a case is a biomedical domain \cite{bioner} with categories such as chemical or disease. Finally, annotations could be inferred from an extensive knowledge base such as WikiData \cite{wikidata} and used as distantly supervised examples. Nevertheless, such annotations will inevitably contain a lot of noise from incorrect or missing entity mentions.

\begin{figure}
    \centering
    \includegraphics[width=0.8\linewidth]{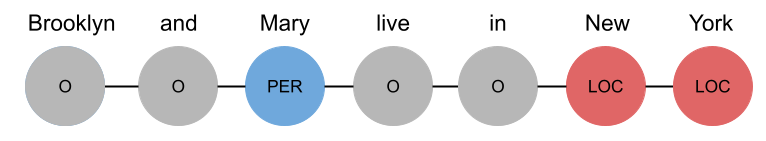}
    \caption{Demonstration of selectively labelled person example in fully labelled dataset of locations. ``Mary`` is selectively labelled, while ``Brooklyn`` token remained labelled as ``Outside of an entity``.}
    \label{fig:partial_example}
\end{figure}

We propose using selective annotation for new entity category introduction to existing corpora. When there is a need for a new category, the training dataset is either populated with new examples mentioning entities of this category or some part of it is relabelled so that some existing mentions are labelled correctly. This process yields a partially labelled dataset, meaning that there are examples of correctly annotated entities present. Yet, some entity mentions are incorrectly labelled as ``Outside of an entity`` since, by default, every unlabelled token is considered to not be a part of an entity. Figure \ref{fig:partial_example} depicts a selectively labelled example.
 
Although the task of training on partially annotated datasets is similar to Few-Shot Learning \cite{few-shot-ner}, the restrictions dictated by these tasks on the available annotations are quite different. Even though Few-Shot Learning methods require a lot less labelled data, such methods expect the available examples to be fully annotated. Given these differences, it is unclear how to compare Few-Shot methods with methods of training on partially annotated datasets that are explored later in this work.

\section{Related Work}

The problem of training on partially annotated datasets is not novel. There have been a number of papers that explored various methods of modelling NER partial annotations (Fig. \ref{annot:partial}) by modifying the CRF loss function to account for the label uncertainty at non-entity tokens \cite{partial-bioner, distant-reinforcement-partial, bde}. The idea of specific modelling was introduced by \cite{fernandes-partial} and further expanded in later works.

Another complementary approach to partial annotations modelling is dataset ``denoising``---detection and correction of various annotation mistakes. For example, many methods dealing with automatically annotated training data (or distantly supervised) employ strategies that can be applied to partially annotated datasets. This is due to a specific process of distant label generation for NER that usually uses external knowledge bases and dictionary matching to annotate unlabelled data. As shown in \cite{bond, pu-ner, distant-reinforcement-partial}, such a process usually yields annotations with high precision and low recall compared to ground truth. In other words, most of the annotations are correct, yet not all of the named entities are labelled. To overcome low recall of the training dataset, self-supervised methods such as \cite{bond, bde} are used to iteratively relabel the training dataset with the models trained on annotations from the previous iterations. The following sections \ref{sec:bond} and \ref{sec:base} will describe the general methodology behind these methods. 

\subsection{BOND}\label{sec:bond}

\begin{figure}[ht]
    \centering
    \begin{subfloat}[Ground truth annotations\label{annot:gold}]{%
        \includegraphics[width=0.8\linewidth]{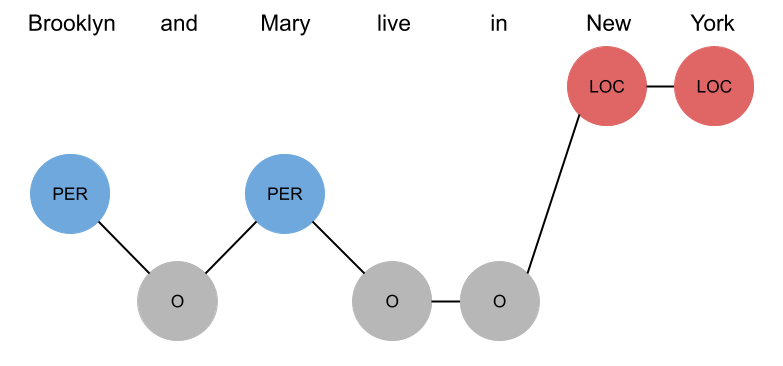}}
    \\
    \end{subfloat}
    \begin{subfloat}[Partial annotations\label{annot:partial}]{%
        \includegraphics[width=0.8\linewidth]{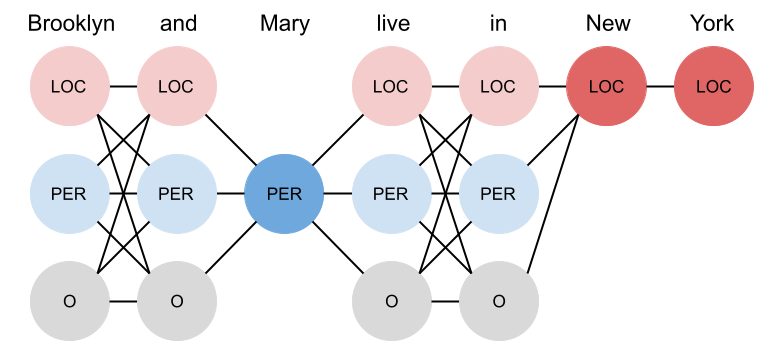}}
    \\
    \end{subfloat} 
    \begin{subfloat}[Teacher annotations\label{annot:teacher}]{%
        \includegraphics[width=0.8\linewidth]{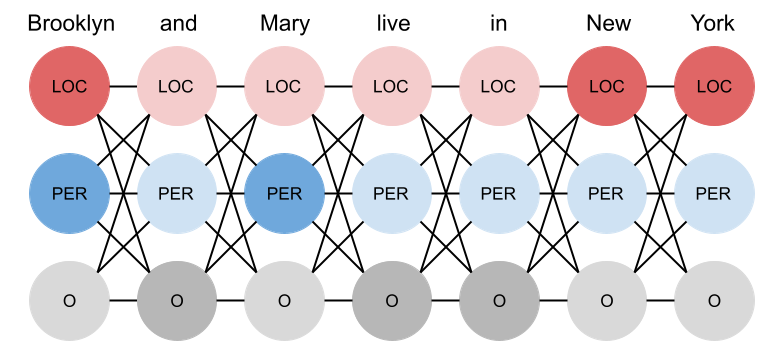}}
    \\
    \end{subfloat}
    \begin{subfloat}[Teacher annotations corrected with partial annotations\label{annot:corrected_teacher}]{%
        \includegraphics[width=0.8\linewidth]{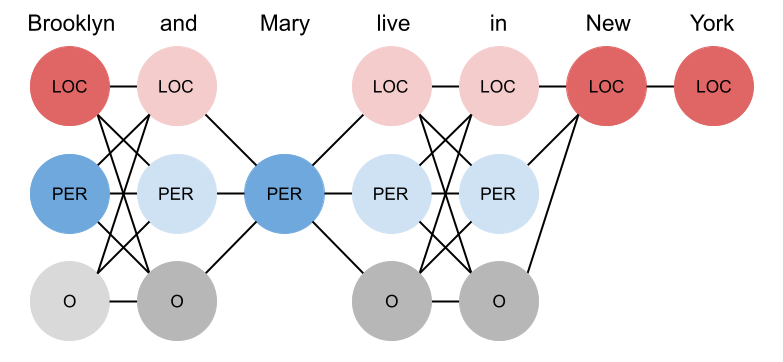}}
    \\
    \end{subfloat} 
    \caption{Demonstration of various types of annotations. Lines indicate the next possible labels, and lighter colours show low-confidence labels. Compared to ground truth annotation (a), partial annotations (b) do not have any confident ``O`` (Outside of an entity) labels and have some of the entity labels missing, e.g. ``Brooklyn`` does not have any confident label. Teacher annotations (c) are expected to be close to ground truth with some noise that comes from using soft targets, e.g. ``Brooklyn`` is assigned ``LOC`` and ``PER`` labels with high confidence.}
    \label{annot} 
\end{figure}

Liang et al. \cite{bond} proposed BOND as a training procedure to finetune BERT on distantly labelled datasets, i.e. automatically annotated by some other model, such as gazetteers crawled from knowledge bases. Their method consists of the following steps:

\begin{enumerate}[Step 1:]
    \item \textit{NER fitting stage}: use early stopping to train the model without overfitting the ``noisy`` annotations.
    \item Define the teacher and student models as copies of the model trained in the previous step.
    \item \textit{Self-training stage}: use the teacher model to obtain the label distributions (See Fig. \ref{annot:teacher}), update the student model w.r.t. to these label distributions as soft targets for several iterations. Then replace the teacher model with the current student model and repeat this step.
\end{enumerate}

Authors of the method state that the third step allows utilising the language understanding encoded in pre-trained BERT weights to correct any missing or incorrect annotations of the entities.

While BOND has shown excellent performance on distantly supervised data, it needs to be clarified how well it is suited for training on partially annotated datasets. Additionally, it is essential to note that this training method does require a fully labelled high-quality validation dataset to detect overfitting during the NER fitting stage and for the model selection in the self-training stage.

\subsection{Base Distribution Estimation (BDE)}\label{sec:base}

Jie et al. \cite{bde} combined partial annotation modelling and self-supervision to correct missing annotations for the partially annotated training dataset. The authors propose using a modification of the CRF decoder that can be trained on label distributions (soft labels) rather than on labels and a specific training procedure:

\begin{enumerate}[Step 1:]
    \item Estimate the ``base distribution`` in cross-validation fashion:
    \begin{enumerate}[1.1]
        \item Partition the training dataset into $k$ equal parts: $K_1, ..., K_k$.
        \item For each $K_i$ train a model on $\{K_1, ..., K_k\} \setminus K_i$ and use it to estimate label distributions $d_i$ for $K_i$ dataset part.
    \end{enumerate}
    \item Train the final model on the whole dataset using the obtained label distributions $d_1, ..., d_k$ as soft targets.
\end{enumerate}

While the authors of this method used an LSTM-based architecture for the model, in practice, such a training procedure can be applied to any model that can be trained with soft targets.

Even though the authors never specifically mentioned it, analyzing the codebase they have provided led us to the conclusion that they might have used a fully labelled validation set to select the best models for the BDE stage and during the main training procedure (Step 2). Moreover, they also utilized an adaptive learning rate schedule based on validation metrics during training.

\section{Our Approach}

Both training procedures described in sections \ref{sec:base} and \ref{sec:bond} can be used to train models on partially annotated datasets since they employ self-supervised methods that can recover missing annotations. Due to the fact that the BDE method can be seen as some preprocessing of target labels, it can be utilized to discover the missing annotations in training data before the main training procedure. Moreover, some noise-resistant model or training method can be used during the base distributions estimation to increase the quality of preprocessed annotations further.

While using BOND to finetune BERT on partially annotated data is possible, the available annotations will not be used efficiently since all annotations will be discarded after the NER fitting stage and replaced with the soft targets of the teacher model. To solve this issue, we implement a modification of the BOND training procedure---GuidedBOND. During the self-training stage, we do not discard the annotated entities $E$ but instead, use them to ``guide`` the model in the right direction by correcting the label distributions $D$ obtained from the teacher model.

Let $C$ be the number of unique labels, $L$---length of the token sequence and $N$---the number of annotated entities. Given the sequence of encoded labels $(l_1, ..., l_L)$, $l_i \in (1, ..., C)$, $E$ can be simplified as a set of non-intersecting continuous label subsequences, indicating which labels are known to be labelled correctly:

\begin{equation}\notag
    \begin{aligned}
        & E = \{ (m_i, n_i) \mid 1 \leq m_i < n_i \leq L + 1, ~ i = 1, ..., N \}
    \end{aligned}
\end{equation}

Teacher model label distributions $D$ are the probability vectors $d_k$ with $p^k_j$ components, $\sum_{j=1}^C p_j^k = 1$. The index $k$ indicates that $d_k$ was obtained for $k$-th token.

\begin{equation}\notag
    \begin{aligned}
        & D = \{ d_k \mid d_k = (p_1^k, p_2^k, ..., p_C^k), ~ k = 1, ... , L \}
    \end{aligned}
\end{equation}

The corrected label distributions $\hat{D}$ are defined equivalently with $\hat{d}_k$ probability vectors that are the same as corresponding $d_k$ vectors if no entity includes $k$-th token, and a one-hot vector of encoded label $l_k$ otherwise. Formally, $\hat{p}^k_j$ component of $\hat{d}_k$ can be defined as follows:

\begin{equation}\notag
    \hat{p}_j^k = \begin{cases}
        \mathds{1}[j = l_k] & \exists (m, n) \in E \mid m \leq k < n \\
        p_j^i & otherwise
    \end{cases}
\end{equation}

See illustrations \ref{annot:teacher} and \ref{annot:corrected_teacher} for reference. We evaluate our approach in \ref{sec:GuidedBOND} section.

We propose using GuidedBOND during the BDE label preprocessing in the following fashion. First, we estimate base distributions by finetuning the BERT model on $k$ folds with the GuidedBOND training procedure. Next, we use these estimated distributions as soft targets to finetune a pre-trained BERT model. We suppose that utilizing GuidedBOND during the BDE preprocessing stage will yield better estimations on missing annotations since, as we show in section \ref{sec:BERT} and \ref{sec:GuidedBOND}, GuidedBOND yields significant improvements when trained on partial annotations compared to both the conventional supervised BERT finetuning approach and BOND.

\section{Experiments}

In our experiments, we use the English version of the CoNLL 2003 dataset \cite{conll}. We train, validate and test all models on the official dataset split. Following the approach in \cite{bde} on simulating the partial annotations, some fraction of the total entities in the training dataset is removed. Entities in the validation dataset remain fully labelled. Following \cite{bond}, the base version of RoBERTa \cite{roberta} was chosen as a pre-trained BERT model. We report the results of training 5 models with different random initializations on the training dataset with 0.05, 0.1, 0.15, 0.2, 0.3, 0.4 and 0.5 entities labelled. Entity sets were chosen randomly without any regard to initial class imbalances and fixed across all runs\footnote{We were not able to fix entity sets for LSTM-CRF baseline since we have used an implementation provided by the authors \cite{bde}.}. Lines in the Figures \labelcref{fig:bert_vs_baseline,fig:guidance,fig:main_results,fig:framework_ablation,fig:bde_ablation} indicate the mean of 5 runs and a lighter coloured area around them demonstrates the standard deviation. We release the code for the experiments as well as the reproduction instructions.\footnote{ \url{https://github.com/ViktorooReps/guided-bond}}

In the following sections, the specific naming convention is used:

\renewcommand{\arraystretch}{1.2}
\begin{table}[h]
    \centering
    \begin{tabular}{c|p{5cm}}
        Method & Meaning \\
        \hline 
        GuidedBOND / BOND & GuidedBOND / BOND training procedure is used to finetune RoBERTa \\
        \hline 
        RoBERTa & RoBERTa is finetuned conventionally \\
        \hline 
        LSTM-CRF & LSTM-CRF is trained conventionally \\
        \hline 
        BDE\textsubscript{M1}+M2 & M2 is trained on base distributions obtained with M1 
    \end{tabular}
\end{table}

\subsection{Baseline} \label{sec:BERT}

\begin{figure}[ht]
    \centering
    \includegraphics[width=0.9\linewidth]{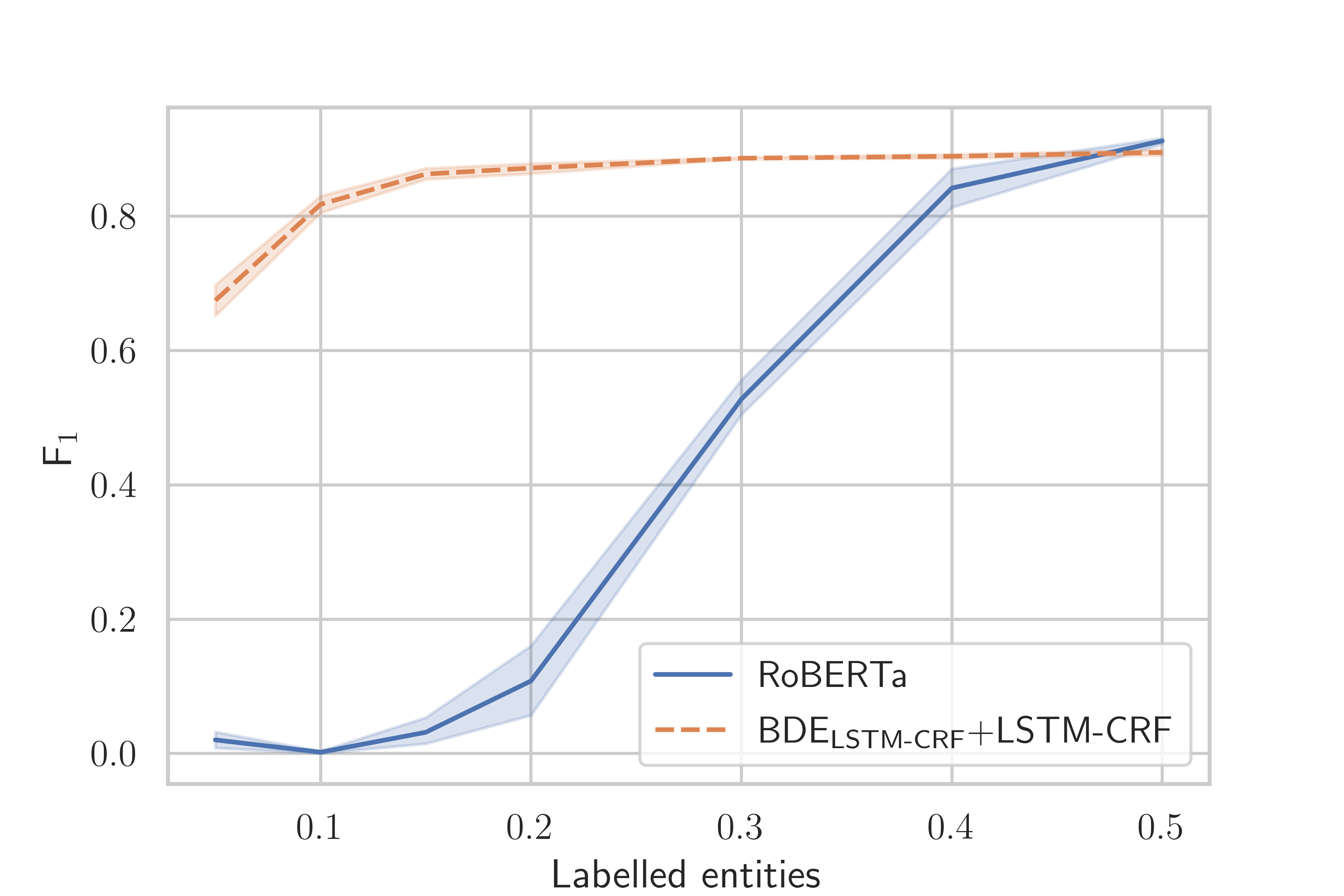}
    \caption{Comparison of conventionally finetuned RoBERTa without BDE with LSTM-CRF model trained with BDE proposed by \cite{bde}. RoBERTa needs half of the total entities to be annotated to reach the performance of the BDE-based approach.}
    \label{fig:bert_vs_baseline}
\end{figure}

First, we confirm that finetuning RoBERTa on partially annotated corpora does, in fact, produce results that are worse than the existent BDE\textsubscript{LSTM-CRF}+LSTM-CRF approach. As demonstrated in Figure \ref{fig:bert_vs_baseline}, more than 50\% of the missing entity annotations in the training data drastically deteriorate the performance of the conventionally finetuned RoBERTa.  

\subsection{Effect of guidance} \label{sec:GuidedBOND}

\begin{figure}[ht]
    \centering
    \includegraphics[width=0.9\linewidth]{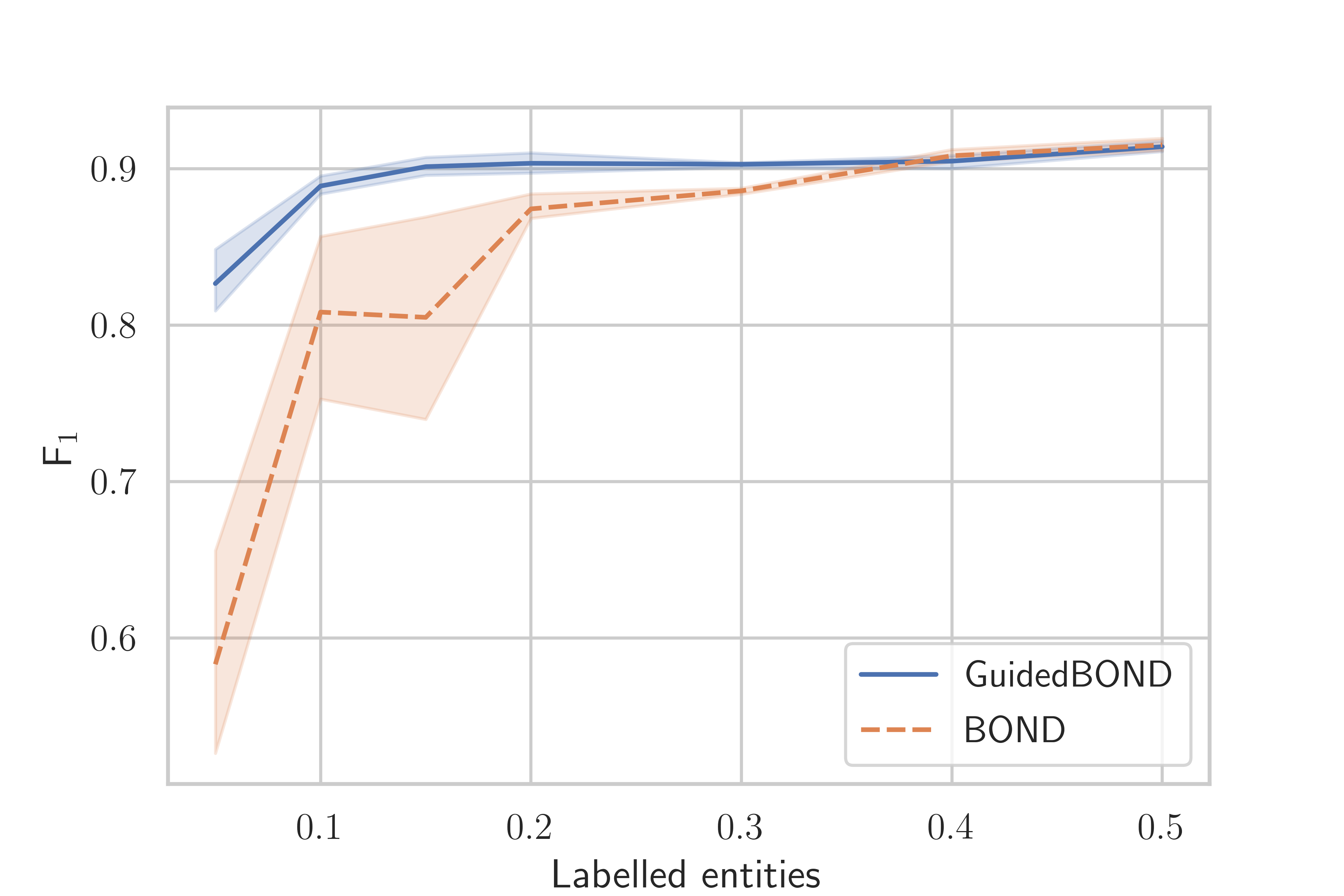}
    \caption{Demonstration of the significance of guidance on BOND training procedure. The guidance on lower fractions of annotated entities makes the finetuning process more stable.}
    \label{fig:guidance}
\end{figure}

Next, we assess the effectiveness of guidance for the BOND training procedure. The experiments illustrated in Figure \ref{fig:guidance} not only confirm that guidance during the self-training stage does help to converge to a better model but also show that on lower fractions of labelled entities, the training process becomes a lot more stable with respect to initialization of the model.

\subsection{Combined approach} \label{sec:BDE+GuidedBOND}

\begin{figure}[ht]
    \centering
    \includegraphics[width=0.9\linewidth]{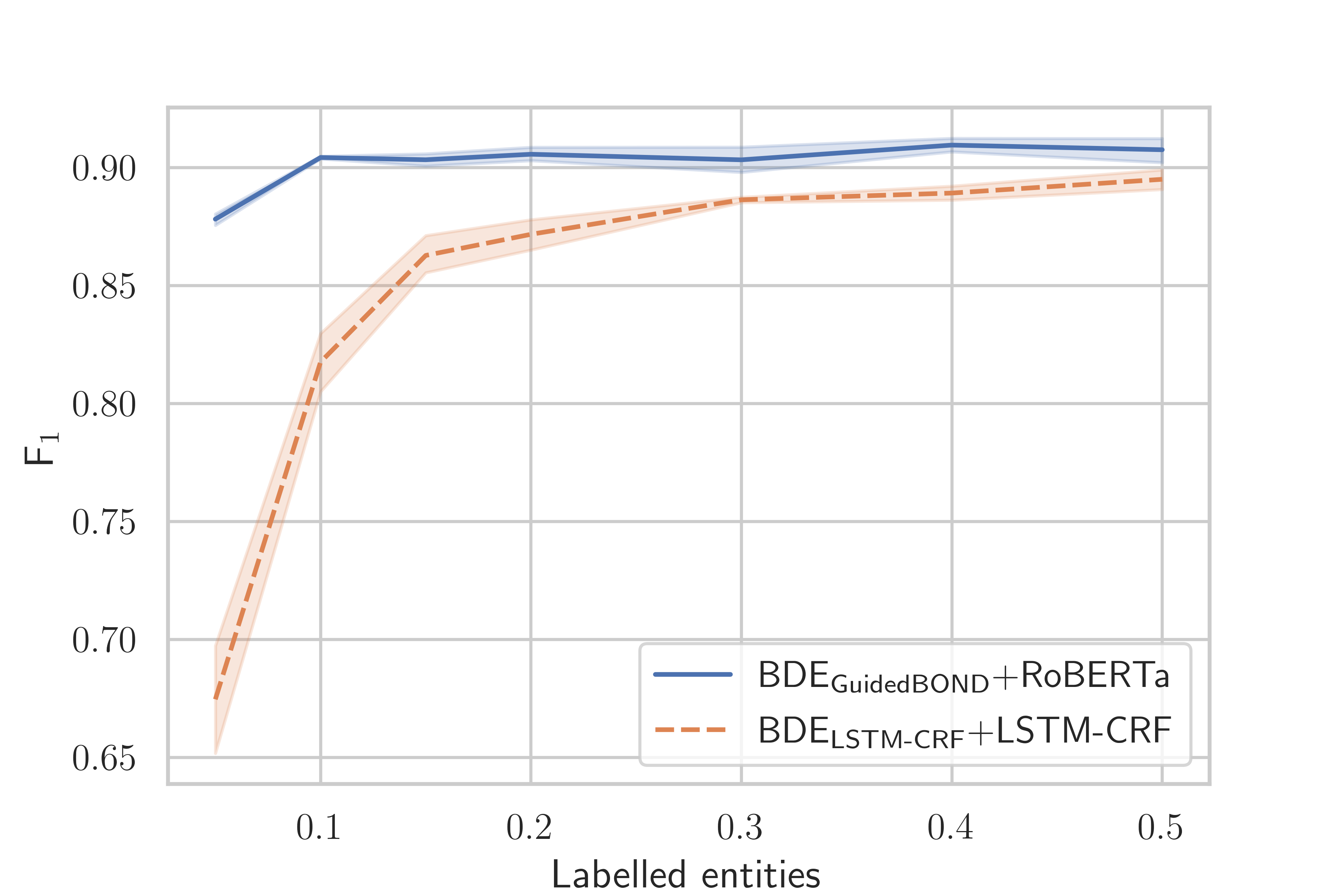}
    \caption{Comparison of RoBERTa finetuned on BDE obtained with GuidedBOND training procedure with LSTM-CRF model trained on BDE proposed by \cite{bde}. Our model is better overall, with the most significant improvements on the lower fractions of annotated entities.}
    \label{fig:main_results}
\end{figure}

Finally, we compare our approach---estimation of base distributions with GuidedBOND training procedure with the LSTM-CRF baseline (see Fig. \ref{fig:main_results}). Unsurprisingly---since BERT was shown to outperform LSTM-based approaches \cite{bert}---our approach yields modest improvements when trained on datasets with higher fractions of labelled entities. Nonetheless, on lower fractions, the improvements become a lot more significant. Our approach reaches a 0.9 F$_1$ score compared to the baseline with close to 5 times fewer annotated entities.

\section{Ablation Study} \label{sec:ablation}

In this section we will evaluate an impact of the different training stages on the final model performance. 

\subsection{Main training procedures}

\begin{figure}[ht]
    \centering
    \includegraphics[width=0.9\linewidth]{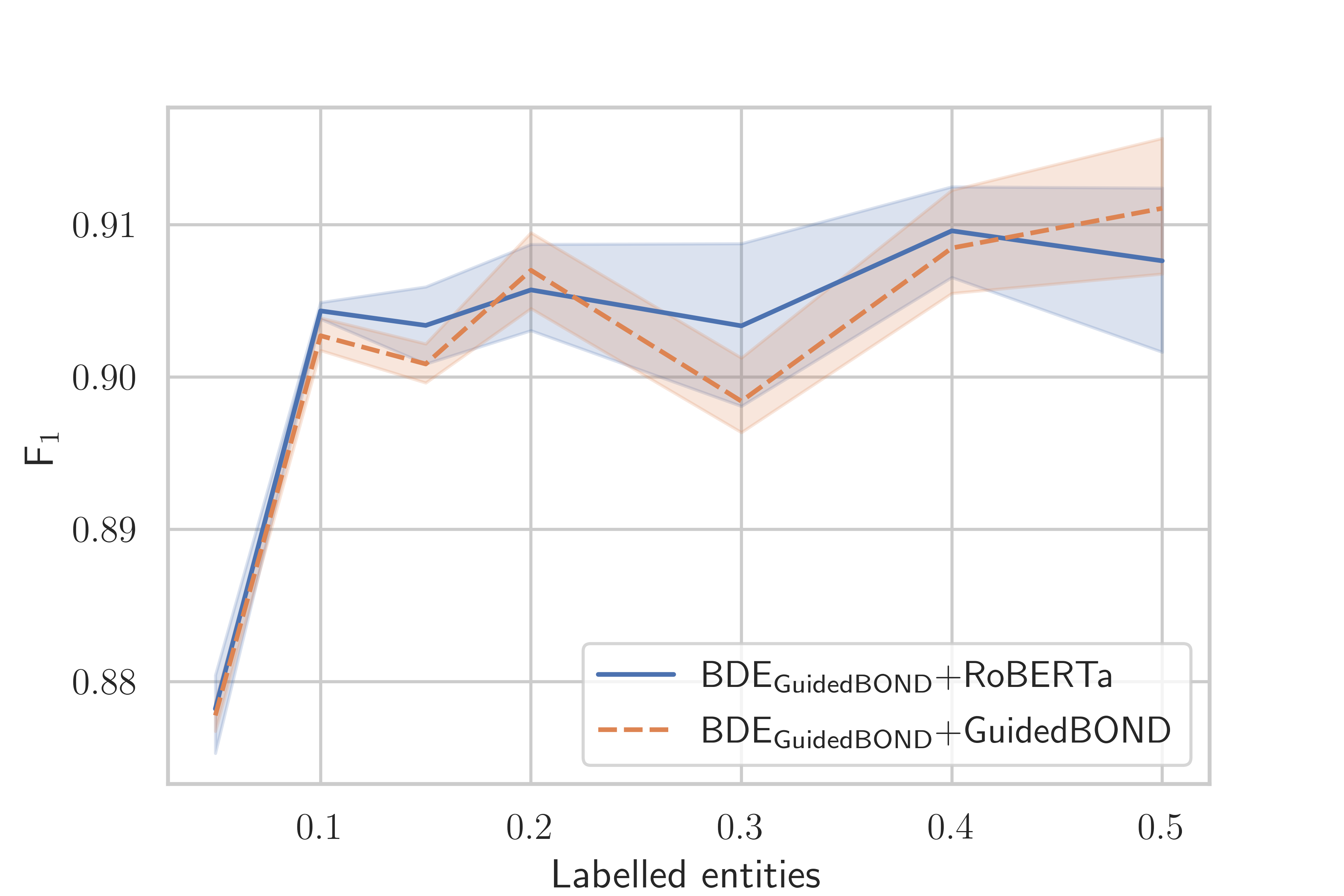}
    \caption{Comparison of different frameworks for training on base distributions. GuidedBOND does not show any significant improvements over the simple supervised approach.}
    \label{fig:framework_ablation}
\end{figure}

Due to the fact that annotations preprocessed with BDE can be considered as distant soft labels, and since BOND was originally designed for training with distant supervision, we attempt to apply GuidedBOND to train models on estimated base distributions (see the results in Fig. \ref{fig:framework_ablation}). We report no improvements compared to the supervised approach. The reason may be that the increase in F$_1$ score during the self-training stage is mainly due to the discovery of the missing annotations, and most of the missing annotations (that are recoverable by self-supervision) are recovered during the BDE stage.

\subsection{Annotations preprocessing}

\begin{figure}[ht]
    \centering
    \includegraphics[width=0.9\linewidth]{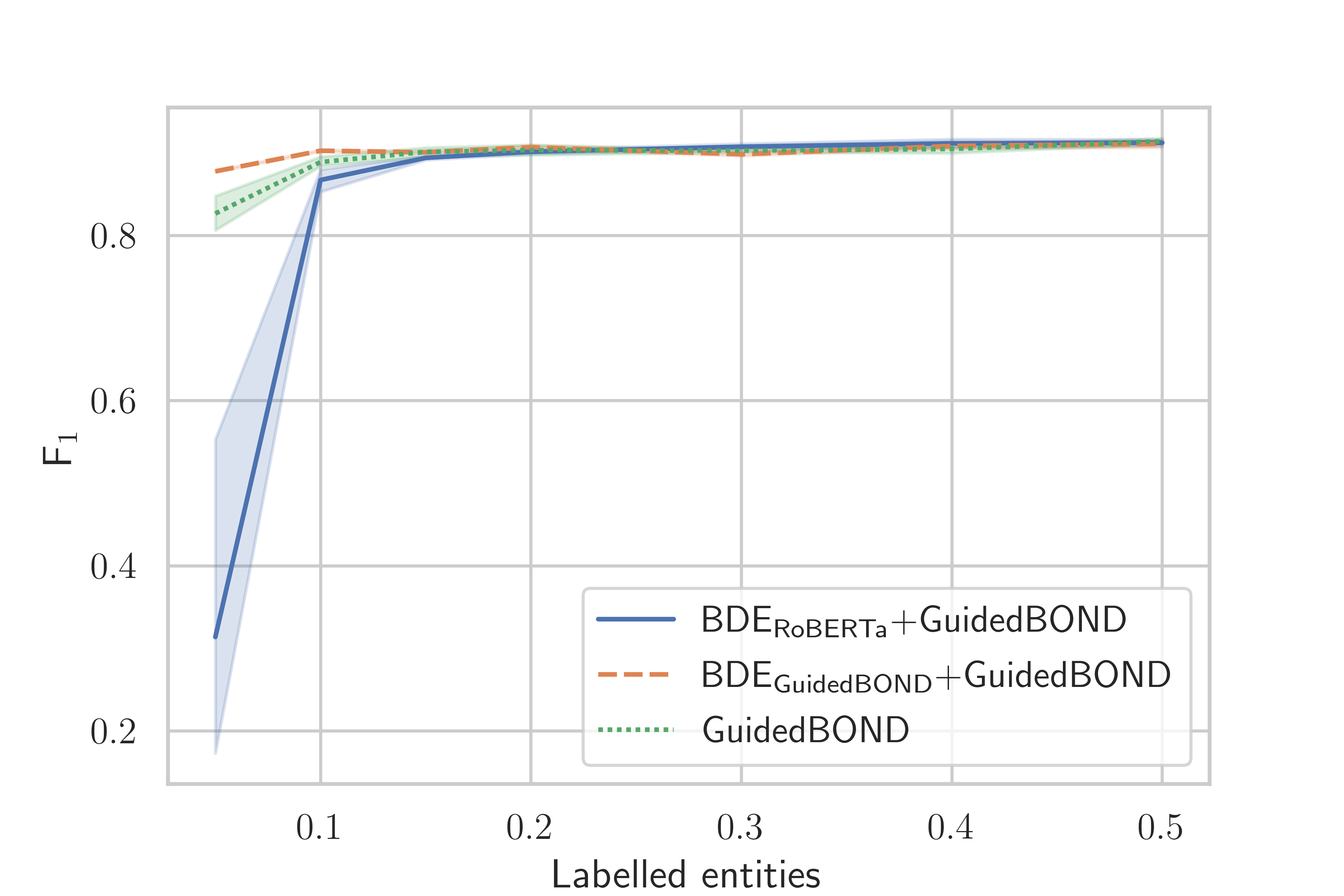}
    \caption{Comparison of different ways to obtain the base distributions. Out of three approaches GuidedBOND yields the most stable results.}
    \label{fig:bde_ablation}
\end{figure}

To evaluate whether using GuidedBOND during the BDE stage (and the stage itself) is necessary, we conduct an experiment by training RoBERTa with GuidedBOND on base distributions obtained from conventionally finetuned RoBERTa and RoBERTa finetuned with GuidedBOND. The comparison between these models and the model trained without the BDE stage is demonstrated in Figure \ref{fig:bde_ablation}. We conclude that BDE yields no improvements except for the cases with extremely low fractions of entities labelled, where, interestingly enough, a fully supervised approach of estimating base distribution leads to exceptionally poor performance. 

\section{Conclusion and Future Work} \label{conclusion}

We have proposed an approach based on a self-supervised training procedure and preprocessing of the available annotations for finetuning BERT models on partially annotated data. Our experiments have shown that our approach outperforms the previous method of dealing with partial annotations, with substantial improvements on lower fractions of annotated entities.

Nevertheless, our approach, as well as the approaches it is based on, relies on a fully annotated development dataset for the model selection and for determining the learning rate schedule. Due to the reasons discussed in the previous sections, it is often not feasible to collect such a dataset under real conditions. This complicates the application of our method to real-world tasks.

Our future work will be focused on researching the approaches to evaluation on partially labelled datasets, as well as on the application of our method to training on several combined datasets with different sets of entity categories.

\printbibliography

\end{document}